\documentclass[runningheads]{llncs}
\usepackage[T1]{fontenc}
\usepackage{graphicx}
\usepackage{amssymb}
\usepackage[percent]{overpic}
\usepackage{dcolumn}
\usepackage{booktabs}
\begin{document}
\title{BS3D: Building-scale 3D Reconstruction from RGB-D Images}
%
%

\author{Janne Mustaniemi\inst{1} \and
Juho Kannala\inst{2} \and
Esa Rahtu\inst{3} \and \\
Li Liu\inst{1} \and
Janne Heikkil\"a\inst{1}}
\authorrunning{Mustaniemi et al.}
\institute{Center for Machine Vision and Signal Analysis, University of Oulu,
 Finland \and
Department of Computer Science, Aalto University, Finland \and
Tampere University, Finland\\
\email{janne.mustaniemi@oulu.fi}}


%
\maketitle              
\begin{abstract}
Various datasets have been proposed for simultaneous localization and mapping (SLAM) and related problems. Existing datasets often include small environments, have incomplete ground truth, or lack important sensor data, such as depth and infrared images. We propose an easy-to-use framework for acquiring building-scale 3D reconstruction using a consumer depth camera. Unlike complex and expensive acquisition setups, our system enables crowd-sourcing, which can greatly benefit data-hungry algorithms. Compared to similar systems, we utilize raw depth maps for odometry computation and loop closure refinement which results in better reconstructions. We acquire a building-scale 3D dataset (BS3D) and demonstrate its value by training an improved monocular depth estimation model. As a unique experiment, we benchmark visual-inertial odometry methods using both color and active infrared images.


\end{abstract}
%
%
%
\section{Introduction}
Simultaneous localization and mapping (SLAM) is an essential component in robot navigation, virtual reality (VR), and augmented reality (AR) systems. Various datasets and benchmarks have been proposed for SLAM \cite{geiger2013vision,sturm12iros,wang2020tartanair} and related problems, including visual-intertial odometry \cite{schubert2018tum,cortes2018advio}, camera re-localization \cite{sattler2018benchmarking,shotton2013scene,kendall2015posenet}, and depth estimation \cite{Silberman:ECCV12,song2015sun}. Currently, there exists only a few building-scale SLAM datasets \cite{sarlin2022lamar} that include ground truth camera poses and dense 3D geometry. Such datasets enable, for example, evaluation of algorithms needed in large-scale AR applications.

The lack of building-scale SLAM datasets is explained by the difficulty of acquiring ground truth data. Some have utilized a high-end LiDAR for obtaining 3D geometry of the environment \cite{ramakrishnan2021habitat,burri2016euroc,chang2017matterport3d,sarlin2022lamar}. Ground truth camera poses may be acquired using a motion capture (MoCap) system when the environment is small enough \cite{sturm12iros,wasenmuller2016corbs}. The high cost of equipment, complex sensor setup, and slow capturing process make these approaches less attractive and inconvenient for crowd-sourced data collection.

An alternative is to perform 3D reconstruction using a monocular, stereo, or depth camera. Consumer RGB-D cameras, in particular, are interesting because of their relatively good accuracy, fast acquisition speed, low-cost, and effectiveness in textureless environments. 
RGB-D cameras have been used to collect datasets for depth estimation \cite{Silberman:ECCV12,song2015sun}, scene understanding \cite{dai2017scannet}, and camera re-localization \cite{shotton2013scene,valentin2016learning}, among other tasks. The problem is that existing RGB-D reconstruction systems (e.g. \cite{newcombe2011kinectfusion,dai2017bundlefusion,choi2015robust}) are limited to room-scale and apartment-scale environments.

Synthetic SLAM datasets have also been proposed \cite{mccormac2017scenenet,wang2020tartanair,saeedi2019characterizing} that include perfect ground truth. The challenge is that data such as time-of-flight (ToF) depth maps and infrared images are difficult to synthesize realistically. Consequently, training and evaluation done using synthetic data may not reflect algorithm's real-world performance. To address the domain gap problem, algorithms are often fine-tuned using real data.

\begin{figure}
  \centering
  \begin{overpic}[width=1.00\textwidth]{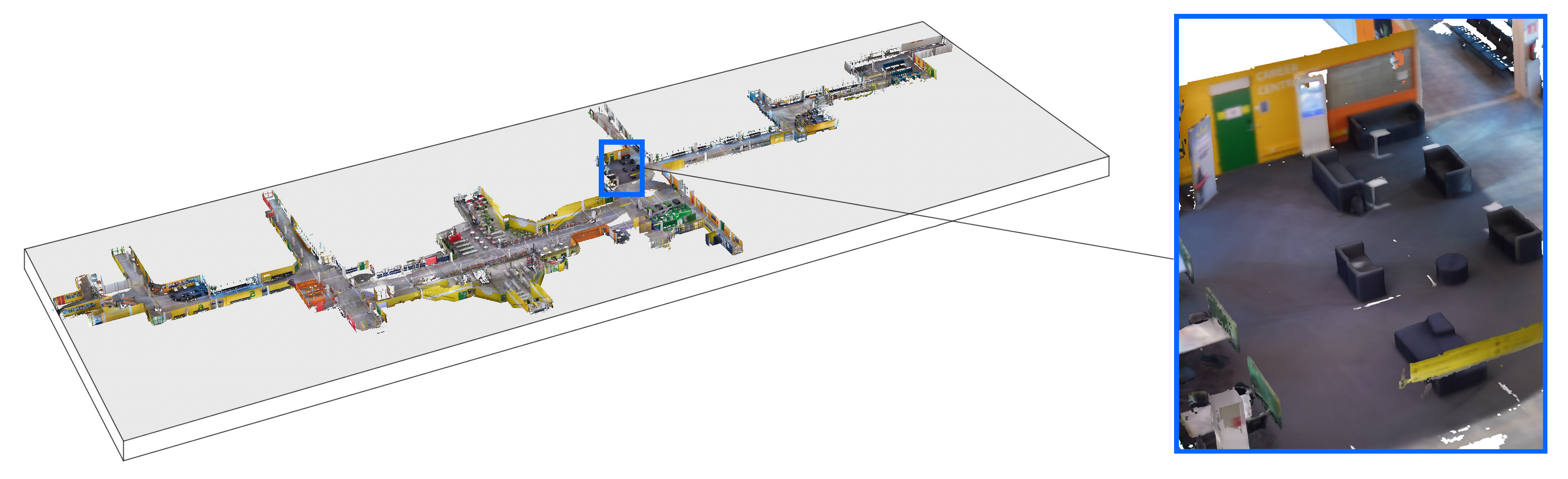}
  \put (26.5,24.0) {\scriptsize 240 m}
  \put (64.3,26.0) {\scriptsize 80 m}
  \put (68,16.5) {\scriptsize 8 m}
  \put (82,30.8) {Zoomed}
  \end{overpic}
  \caption{Building-scale 3D reconstruction (4300 m2) obtained using an RGB-D camera and the proposed framework. The magnified area (90 m2) is larger than any reconstruction in the ScanNet dataset \cite{dai2017scannet}.}
  \label{fig:teaser}
\end{figure}
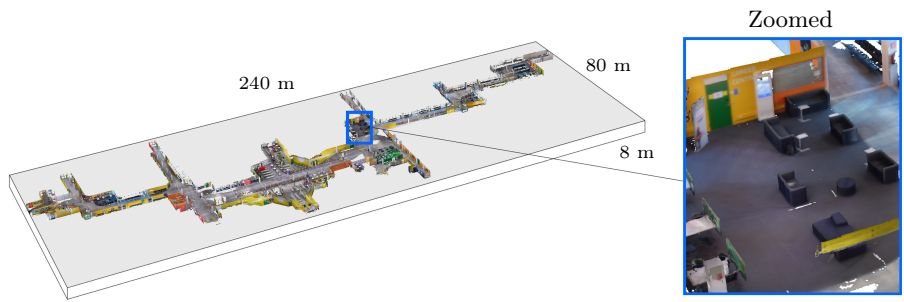

We propose a framework to create building-scale 3D reconstructions using a consumer depth camera (Azure Kinect). Unlike existing approaches, we register color images and depth maps using color-to-depth (C2D) strategy. This allows us to directly utilize the raw depth maps captured by the wide field-of-view (FoV) infrared camera. Coupled with an open-source SLAM library \cite{labbe2019rtab}, we acquire a building-scale 3D vision dataset (BS3D) that is considerably larger than similar datasets as shown in Fig.~\ref{fig:teaser}. The BS3D dataset includes 392k synchronized color images, depth maps and infrared images, inertial measurements, camera poses, enhanced depth maps, surface reconstructions, and laser scans. Our framework will be released for the public to enable fast, easy and affordable indoor 3D reconstruction.

%
%
\section{Related work}
This section introduces commonly used RGB-D SLAM datasets and corresponding data acquisition processes. A summary of the datasets is provided in Table \ref{tab:datasets}. As there exist countless SLAM datasets, the scope is restricted to real-world indoor scenarios. We leave out datasets focusing on aerial scenarios (e.g. EuRoC MAV \cite{burri2016euroc}) and autonomous driving (e.g. KITTI \cite{geiger2013vision}). We also omit RGB-D datasets captured with a stationary scanner (e.g. Matterport3D \cite{chang2017matterport3d}) as they cannot be used for SLAM evaluation. Synthetic datasets, such as SceneNet RGB-D \cite{mccormac2017scenenet}, TartanAir \cite{wang2020tartanair}, and ICL \cite{saeedi2019characterizing} are also omitted.

ADVIO \cite{cortes2018advio} dataset is a realistic visual-inertial odometry benchmark that includes building-scale environments. Ground truth trajectory is computed using an inertial navigation system (INS) together with manual location fixes. The main limitation of the dataset is that it does not come with ground truth 3D geometry. LaMAR \cite{sarlin2022lamar} is a large-scale SLAM benchmark that utilizes high-end mapping platforms (NavVis M6 trolley and VLX backpack) for ground truth generation. Although the capturing setup includes a variety of devices (e.g. HoloLens2 and iPad Pro), it does not include a dedicated RGB-D camera.

OpenLORIS-Scene \cite{shi2020we} focuses on the lifelong SLAM scenario where environments are dynamic and changing, similar to LaMAR \cite{sarlin2022lamar}. The data is collected over an extended period of time using wheeled robots equipped with various sensors, including RGB-D, stereo, IMU, wheel odometry, and LiDAR. Ground truth poses are acquired using an external motion capture (MoCap) system, or with a 2D laser SLAM method. The dataset is not ideal for handheld SLAM evaluation because of the limited motion patterns of a ground robot. 

TUM RGB-D SLAM \cite{sturm12iros} is one of the most popular SLAM datasets. The RGB-D images are acquired using a consumer depth camera Microsoft Kinect v1. Ground truth trajectory is incomplete because the MoCap system can only cover a small part of the environment. CoRBS \cite{wasenmuller2016corbs} consists of four room-scale environments. It also utilizes MoCap for acquiring ground truth trajectories for Microsoft Kinect v2. Unlike \cite{sturm12iros}, CoRBS provides ground truth 3D geometry acquired using a laser scanner. The data also includes infrared images, but not inertial measurements, unlike our dataset.

7-Scenes \cite{shotton2013scene} and 12-Scenes \cite{valentin2016learning} are commonly used for evaluating camera localization. 7-Scenes includes seven scenes captured using Kinect v1. KinectFusion \cite{newcombe2011kinectfusion} is used to obtain ground truth poses and dense 3D models from the RGB-D images. 12-Scenes consists of multiple rooms captured using the Structure.io depth sensor and iPad color camera. The reconstructions are larger compared to 7-Scenes, about 37 m\textsuperscript{3} on average. They are acquired using the VoxelHashing framework \cite{niessner2013real}, an alternative to KinectFusion with better scalability.

ScanNet \cite{dai2017scannet} is an RGB-D dataset containing 2.5M views acquired in 707 distinct spaces. It includes estimated calibration parameters, camera poses, 3D surface reconstructions, textured meshes, and object-level semantic segmentations. 
The hardware consists of a Structure.io depth sensor attached to a tablet computer. Pose estimation is done using BundleFusion \cite{dai2017bundlefusion}, after which volumetric integration is performed through VoxelHashing \cite{niessner2013real}. 

Sun3D \cite{xiao2013sun3d} is a large RGB-D database with camera poses, point clouds, object labels, and refined depth maps. The reconstruction process is based on  structure from motion (SfM) where manual object annotations are utilized to reduce drift and loop-closure failures. Refined depth maps are obtained via volumetric fusion similar to KinectFusion \cite{newcombe2011kinectfusion}. We emphasize that ScanNet \cite{dai2017scannet} and Sun3D \cite{xiao2013sun3d} reconstructions are considerably smaller and have lower quality than those provided in our dataset. Unlike \cite{sarlin2022lamar,shi2020we,sturm12iros}, our system also does not require a complex and expensive capturing setup, or manual annotation \cite{cortes2018advio,xiao2013sun3d}.

\begin{table}
\caption{List of indoor RGB-D SLAM datasets. The BS3D acquisition setup does not require high-end LiDARs \cite{wasenmuller2016corbs,shi2020we,sarlin2022lamar}, MoCap systems \cite{wasenmuller2016corbs,shi2020we,sturm2012benchmark}, or manual annotation \cite{xiao2013sun3d,cortes2018advio}. BS3D is building-scale, unlike \cite{shotton2013scene,sturm2012benchmark,dai2017scannet,wasenmuller2016corbs,valentin2016learning,xiao2013sun3d}. Note that ADVIO \cite{cortes2018advio} and LaMAR \cite{sarlin2022lamar} do not have a dedicated depth camera.}
\label{tab:datasets} 
\setlength{\tabcolsep}{5pt}
\begin{tabular}{@{\hspace{\tabcolsep}}l|lccccc}
\hline\noalign{\smallskip}
Dataset & Scale & Depth & IMU & IR & Ground truth \\
\noalign{\smallskip}\hline\noalign{\smallskip}
7-Scenes \cite{shotton2013scene} & room & Kinect v1 & - & - & RGBD-recons. \\
TUM RGBD \cite{sturm2012benchmark} & room & Kinect v1 & \checkmark & - & MoCap\\
ScanNet \cite{dai2017scannet} & room & Structure.io & \checkmark & - & RGBD-recons. \\
CoRBS \cite{wasenmuller2016corbs} & room & Kinect v2 & - & \checkmark & MoCap+LiDAR\\
12-Scenes \cite{valentin2016learning} & apartment & Structure.io  & - & - & RGBD-recons. \\
Sun3D \cite{xiao2013sun3d} & apartment & Xtion Pro Live & - & - & RGBD+manual \\
OpenLORIS \cite{shi2020we} & building & RS-D435i & \checkmark & - & MoCap+LiDAR \\
ADVIO \cite{cortes2018advio} & building & Tango & \checkmark & - & INS+manual \\
LaMAR \cite{sarlin2022lamar} & building & HoloLens2 & \checkmark & \checkmark  & LiDAR+VIO+SfM \\
\noalign{\smallskip}
\hline\noalign{\smallskip}
\textbf{BS3D (ours)} & building & Azure Kinect & \checkmark & \checkmark & RGBD-recons. \\
\noalign{\smallskip}\hline
\end{tabular}
\end{table}

%
%
\section{Reconstruction framework}
\label{sec:framework}
\begin{figure}
  \centering
  \includegraphics[width=1.00\textwidth]{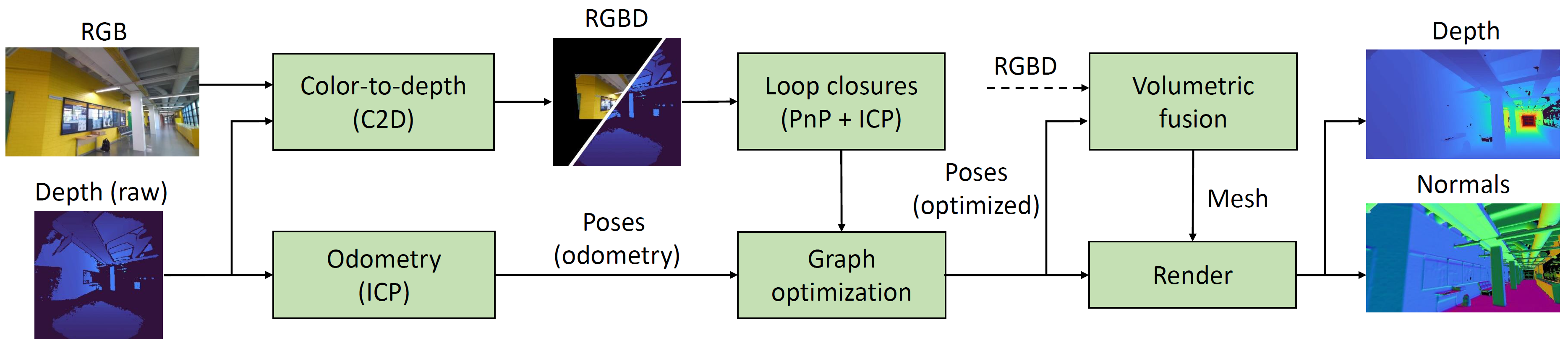}
  \caption{Overview of the RGB-D reconstruction system.}.
  \label{fig:system}
\end{figure}

In this section, we introduce the RGB-D reconstruction framework shown in Fig. \ref{fig:system}. The framework produces accurate 3D reconstructions of building-scale environments using low-cost hardware. The system is fully automatic and robust against poor lighting conditions and fast motions. Color images are only used for loop closure detection as they are susceptible to motion blur and rolling shutter distortion. Raw depth maps enable accurate odometry and the refinement of loop closure transformations.

\subsection{Hardware}
\label{sec:hardware}
Data is captured using the Azure Kinect depth camera, which is well-suited for crowd-sourcing due to its popularity and affordability. We capture synchronized depth, color, and infrared images at 30 Hz using the official recorder application running on a laptop computer. We use the wide FoV mode of the infrared camera with 2x2 binning to extend the Z-range. The resolution of raw depth maps and IR images is 512 x 512 pixels. Auto-exposure is enabled when capturing color images at the resolution of 720 x 1280 pixels. We also record accelerometer and gyroscope readings at 1.6 kHz.

\subsection{Color-to-depth alignment}
\label{sec:c2d_alignment}
Most RGB-D reconstruction systems expect that color images and depth maps have been spatially and temporally aligned. Modern depth cameras typically produce temporally synchronized images so the main concern is the spatial alignment. Conventionally, raw depth maps are transformed to the coordinate system of the color camera, which we refer to as the depth-to-color (D2C) alignment. 

In the case of Azure Kinect, the color camera's FoV is much narrower (90 x 59 degrees) compared to the infrared camera (120 x 120 degrees). Thus, the D2C alignment would not take advantage of the infrared camera's wide FoV because depth maps would be heavily cropped. Moreover, the D2C alignment might introduce artefacts to the raw depth maps.

We propose an alternative called color-to-depth (C2D) alignment where color images are transformed instead. In the experiments, we show that this drastically improves the quality of the reconstructions. The main challenge of C2D is that it requires a fully dense depth map. Fortunately, a reasonably good alignment can be achieved even with a low quality depth map. This is because the baseline between the cameras is narrow and missing depths often appear in areas that are far away from the camera.

For the C2D alignment, we first perform depth inpainting using linear interpolation. Then, the color image is transformed to the raw depth frame. To keep as much of the color information as possible, the output resolution will be higher (1024 x 1024 pixels) compared to the raw depth maps . After that, holes in the color image due to occlusions are inpainted using the OpenCV library's implementation of \cite{telea2004image}. We note that minor artefacts in the aligned color images will have little impact on the SIFT-based loop closure detection.

\subsection{RGB-D Mapping}
\label{sec:single-session}
We process the RGB-D sequences using an open-source SLAM library called RTAB-Map \cite{labbe2019rtab}. Odometry is computed from the raw depth maps using the point-to-plane variant of the iterative closest point (ICP) algorithm \cite{Pomerleau12comp}. We use the scan-to-map odometry strategy \cite{labbe2019rtab} where incoming frames are registered against a point cloud map created from past keyframes. The wide FoV ensures that ICP-odometry rarely fails, but in case it does, a new map is initialized.

Loop closure detection is needed for drift correction and merging of individual maps. For this purpose, SIFT features are extracted from the valid area of the aligned color images. Loop closures are detected using the bag-of-words approach \cite{labbe2013appearance}, and transformations are estimated using the Perspective-n-Point RANSAC algorithm and refined using ICP \cite{Pomerleau12comp}. Graph optimization is done using the GTSAM library \cite{dellaert2012factor} and Gauss-Newton algorithm. 

RTAB-Map supports multi-session mapping which is a necessary feature when reconstructing building-scale environments. It is not practical to collect possibly hours of data at once. Furthermore, having the ability to later update and expand the map is a useful feature. In practise, individual sequences are first processed separately, followed by multi-session mapping. The sessions are merged by finding loop closures and by performing graph optimization. The input is a sequence of keyframes along with odometry poses and SIFT features computed during single-session mapping. The sessions are processed in such order that there is at least some overlap between the current session and the global map build so far.


\subsection{Surface reconstruction}
It is often useful to have a 3D surface representation of the environment. There exists many classical \cite{kazhdan2006poisson,newcombe2011kinectfusion} and learning-based \cite{weder2021neuralfusion,azinovic2022neural} surface reconstruction approaches. Methods that utilize deep neural networks, such as NeuralFusion \cite{weder2021neuralfusion}, have produced impressive results on the task of depth map fusion. Neural radiance fields (NeRFs) have also been adapted to RGB-D imagery \cite{azinovic2022neural} showing good performance. We did not use learning-based approaches in this work because they are limited to small scenes, at least for the time being. Moreover, scene-specific learning \cite{azinovic2022neural} takes several hours even with powerful hardware.

Surface reconstruction is done in segments due to the large scale of the environment and the vast number of frames. To that end, we first create a point cloud from downsampled raw depth maps. Every point includes a view index along with 3D coordinates. The point cloud is partitioned into manageable segments using the K-means algorithm. A mesh is created for each segment using the scalable TSDF fusion implementation \cite{zhou2018open3d} that is based on \cite{curless1996volumetric,newcombe2011kinectfusion}. It uses a hierarchical hashing structure to support large scenes.

%
\section{BS3D dataset}
\begin{figure}
  \centering
  \begin{overpic}[width=1.00\textwidth]{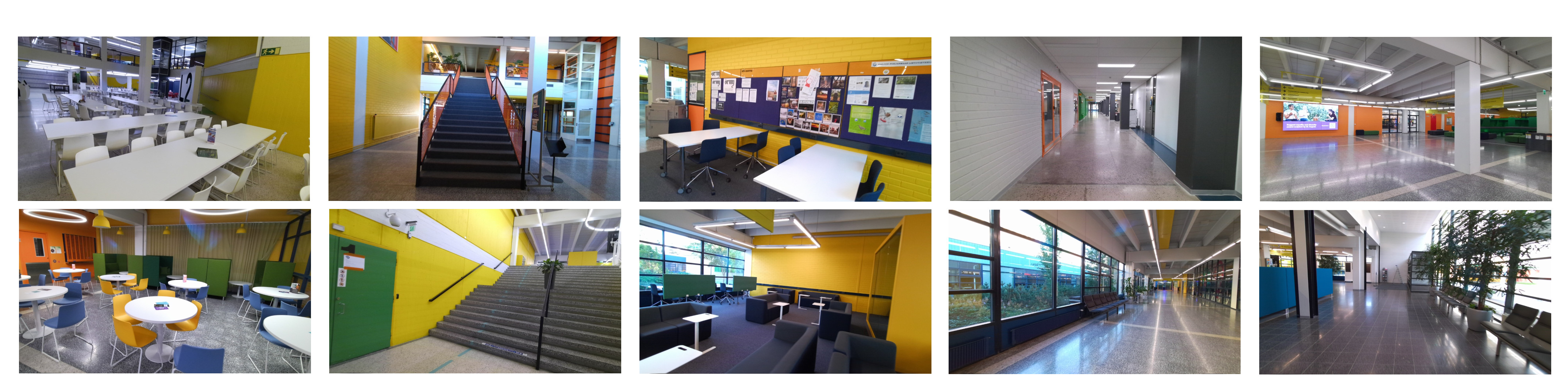}
  \put (6,23.7) {\scriptsize Cafeteria}
  \put (27,23.7) {\scriptsize Stairs}
  \put (47,23.7) {\scriptsize Study}
  \put (65,23.7) {\scriptsize Corridor}
  \put (87,23.7) {\scriptsize Lobby}
  \end{overpic}
  \caption{Example frames from the dataset. Environments are diverse and challenging, including cafeterias, stairs, study areas, corridors, and lobbies.}
  \label{fig:scenes}
\end{figure}
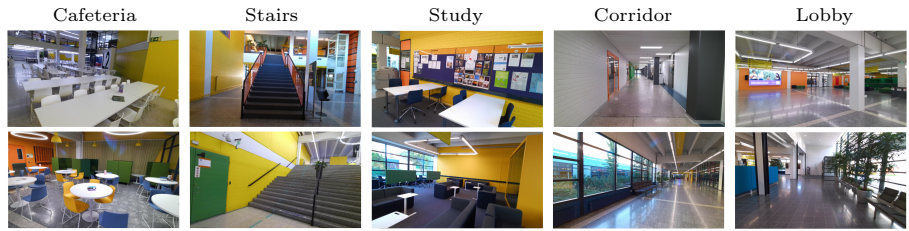

The BS3D dataset was collected at the university campus using Azure Kinect (Section \ref{sec:hardware}). Figure \ref{fig:scenes} shows example frames from the dataset. The collection was done in multiple sessions due to large scale of the environment. The recordings were processed using the framework described in Section \ref{sec:framework}.

\subsection{Dataset features}
The reconstruction shown in Fig. \ref{fig:teaser} consists of 47 overlapping recording sessions. Additional 14 sessions, including 3D laser scans, were recorded for evaluation purposes. Most sessions begin and end at the same location to encourage loop closure detection. The total duration of the sessions is 3 hours and 38 minutes and the combined trajectory length is 6.4 kilometers. The reconstructed floor area is approximately 4300 m$^2$.

\begin{figure}
  \centering
  \begin{overpic}[width=1.00\textwidth]{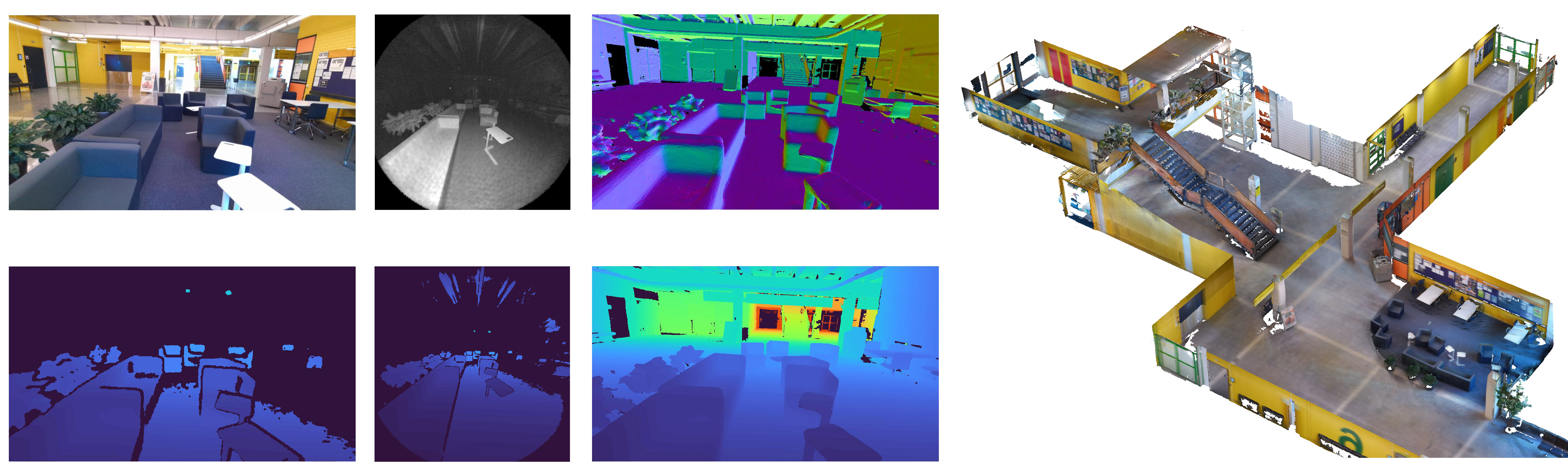}
  \put (9,29.9) {\scriptsize{Color}}
  \put (25.9,29.9) {\scriptsize{Infrared}}
  \put (40,29.9) {\scriptsize{Normals (render)}}
  \put (79,29.9) {\scriptsize{Mesh}}
  \put (9,13.8) {\scriptsize{Depth}}
  \put (23.8,13.8) {\scriptsize{Depth (raw)}}
  \put (41,13.8) {\scriptsize{Depth (render)}}
  \end{overpic}
  \caption{The BS3D dataset includes color and infrared images, depth maps, IMU data, camera parameters, and surface reconstructions. Enhanced depth maps and surface normals are rendered from the mesh.}
  \label{fig:data}
\end{figure}
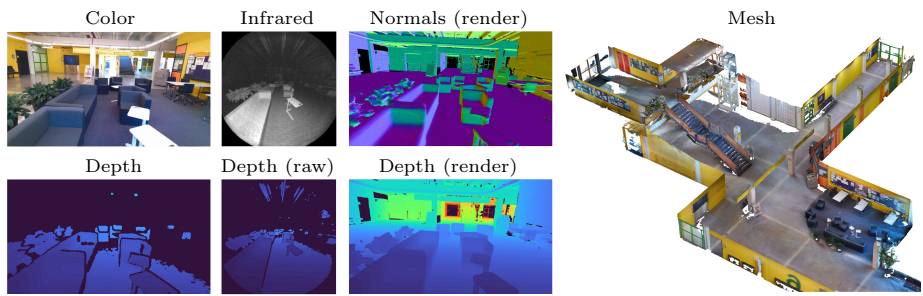

The dataset consists of 392k frames, including color images, raw depth maps, and infrared images. Color images and depth maps are provided in both coordinate frames (color and infrared camera). The images have been undistorted for convenience, but the original recordings are also included. We provide camera poses in a global reference frame for every image. Data also includes inertial measurements, enhanced depth maps and surface normals that have been rendered from the mesh as visualized in Fig.~\ref{fig:data}.




\subsection{Laser scan}
\label{sec:laser_scan}
We utilize the FARO 3D X 130 laser scanner for acquiring ground truth 3D geometry. The scanned area includes a lobby and corridors of different sizes (800 m$^2$). The 28 individual scans were registered using the SCENE software that comes with the laser scanner. Noticeable artefacts, e.g. those caused by mirrors, were manually removed. The laser scan is used to evaluate the reconstruction framework in Section \ref{sec:experiments}. However, this data also enables, for example, training and evaluation of RGB-D surface reconstruction algorithms.

%
%
\section{Experiments}
\label{sec:experiments}
We compare our framework with the state-of-the-art RGB-D reconstruction methods \cite{choi2015robust,dai2017bundlefusion,campos2021orb}. The value of the BS3D dataset is demonstrated by training a recent monocular depth estimation model \cite{Wei2021CVPR}. We also benchmark visual-inertial odometry approaches \cite{geneva2020openvins,von2022dm,campos2021orb} using either color or infrared images to further highlight the unique aspects of the BS3D dataset.

\subsection{Reconstruction framework}
In this experiment, we compare the framework against Redwood \cite{choi2015robust}, BundleFusion \cite{dai2017bundlefusion}, and ORB-SLAM3 \cite{campos2021orb}. RGBD images are provided for \cite{choi2015robust,dai2017bundlefusion,campos2021orb} in the coordinate frame of the color camera. Given the estimated camera poses, we create a point cloud and compare it to the laser scan (Section \ref{sec:laser_scan}). The evaluation metrics include overlap of the point clouds and RMSE of inlier correspondences. Before comparison, we create uniformly sampled point clouds using voxel downsampling (1 cm$^3$ voxel) that computes the centroid of the points in each voxel. The overlap is defined as the ratio of inlier correspondences and the number of ground truth points. A 3D point is considered to be an inlier if the distance to the closest ground truth point is below threshold $\gamma$.

Table \ref{tab:reconstruction} shows the results for environments of different sizes. All methods are able to reconstruct the small environment (35 m$^2$) consisting of 2.8k frames. The differences between the methods become more evident when reconstructing the medium-size environment (160 m$^2$) consisting of 7.3k frames. BundleFusion \cite{dai2017bundlefusion} only produces a partial reconstruction because of odometry failures. The proposed approach gives the most accurate reconstructions as visualized in Fig. \ref{fig:reconstructions}. Note that it is not possible to achieve 100 \% overlap because the depth camera does not observe all parts of the ground truth.

The largest environment (350 m$^2$) consists of 24k frames acquired in four sessions. Redwood \cite{choi2015robust} does not scale to input sequences of this long. ORB-SLAM3 \cite{campos2021orb} frequently loses the odometry in open spaces which leads to incomplete and less accurate reconstructions. Our method suffers the same problem when C2D is disabled. Unreliable odometry is likely due to the color camera's limited FoV, rolling shutter distortion, and motion blur. The C2D alignment improves the accuracy and robustness of ICP-based odometry and loop closures. Without C2D, the frequent odometry failures result in disconnected maps and noticeable drift. We note that the reconstruction in Fig. \ref{fig:teaser} was computed from $\sim$300k frames which is far more than \cite{choi2015robust,dai2017bundlefusion,campos2021orb} can handle.


\begin{table*}
\newcolumntype{.}{D{.}{.}{-1}}
\centering
\setlength{\tabcolsep}{4.0pt}
\caption{Comparison of RGB-D reconstruction methods in small, medium and large-scale environments (from top to bottom). Overlap of the point clouds and inlier RMSE computed for distance thresholds $\gamma$ (mm). Some methods only work in small and/or medium scale environments.}
\begin{tabular}{lllllll}
\toprule
\multicolumn{1}{c}{} &
\multicolumn{2}{c}{$\gamma = 10$ (mm)} &
\multicolumn{2}{c}{$\gamma = 20$ (mm)} &
\multicolumn{2}{c}{$\gamma = 50$ (mm)} \\
\cmidrule(r){2-3}
\cmidrule(r){4-5}
\cmidrule(r){6-7}
\multicolumn{1}{l}{Method} & 
\multicolumn{1}{l}{Overlap $\uparrow$} &
\multicolumn{1}{l}{RMSE $\downarrow$} &
\multicolumn{1}{l}{Overlap $\uparrow$} &
\multicolumn{1}{l}{RMSE $\downarrow$} &
\multicolumn{1}{l}{Overlap $\uparrow$} &
\multicolumn{1}{l}{RMSE $\downarrow$} \\
\cmidrule(r){2-3}
\cmidrule(r){4-5}
\cmidrule(r){6-7}
Redwood \cite{choi2015robust} & 66.5 & 5.6 & 77.9 & 7.6 & 87.1 & 12.6 \\
BundleFusion \cite{dai2017bundlefusion} & 72.1 & 5.5 & 80.8 & 6.9 & 88.3 &  11.7 \\
ORB-SLAM3 \cite{campos2021orb} & 78.2 & 5.3 & 85.2 & \textbf{6.5} & 91.3 & \textbf{10.6} \\
Prop. (w/o C2D)  & 66.8 & 5.7 & 77.8 & 7.5 & 87.0 & 12.7 \\
Proposed & \textbf{78.4} & \textbf{5.2} & \textbf{85.7} & \textbf{6.5} &  \textbf{91.6} & \textbf{10.6} \\
\bottomrule
\rule{0pt}{3ex}Redwood \cite{choi2015robust} & 30.4 & 6.2 & 44.5 &  9.8 & 63.9 &  19.9 \\
BundleFusion \cite{dai2017bundlefusion} & 8.1 & 6.2 & 11.1 & 9.2 & 14.8 &  18.8 \\
ORB-SLAM3 \cite{campos2021orb}  & 44.3 &  6.0 & 57.7 & 8.7 & 71.0 & 16.2 \\
Prop. (w/o C2D)  & 36.5 & 6.1 & 49.2 & 9.0 & 64.3 & 18.3 \\
Proposed & \textbf{54.1} & \textbf{5.7} & \textbf{64.8} & \textbf{7.7} & \textbf{73.2} & \textbf{13.4} \\
\bottomrule
\rule{0pt}{3ex}ORB-SLAM3 \cite{campos2021orb}  & 9.5 & \textbf{6.3} & 14.4 & 9.9 & 20.8 & 20.7 \\
Prop. (w/o C2D)  & 23.1 &  6.7 & 40.6 &  10.9 &  64.7 &  22.4 \\
Proposed & \textbf{34.7} & 6.4 & \textbf{52.7} & \textbf{10.0} & \textbf{75.0} & \textbf{19.8} \\
\bottomrule
\end{tabular}
\label{tab:reconstruction}
\end{table*}

\begin{figure}
  \centering
  \begin{overpic}[width=1.00\textwidth]{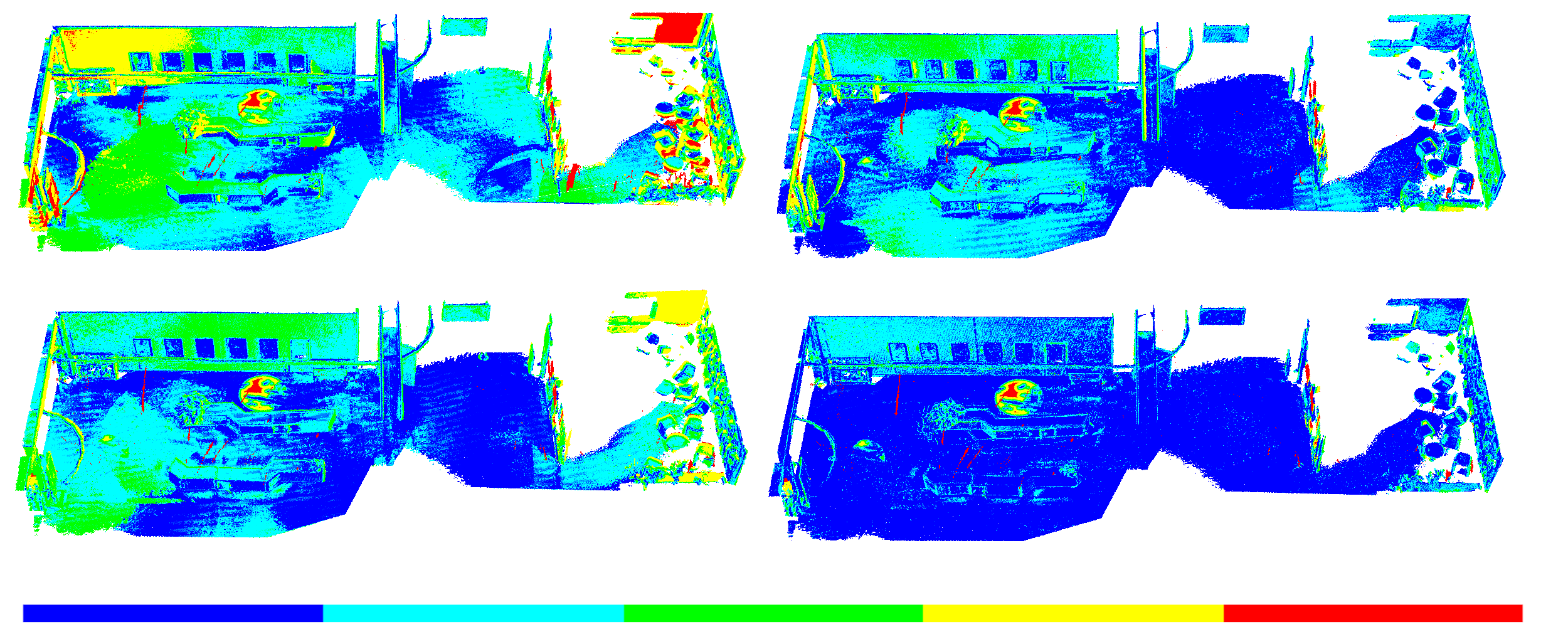}
  \put (63.8,40.0) {ORB-SLAM3 \cite{campos2021orb}}
  \put (67.5,21.2) {Proposed}
  \put (17.0,40.0) {Redwood \cite{choi2015robust}}
  \put (13.5,21.2) {Proposed (w/o C2D)}
  \put (6,2.3) {\scriptsize{$\epsilon < 20$ mm}}
  \put (24.5,2.3) {\scriptsize{$20 \leq \epsilon < 50$}}
  \put (43.5,2.3) {\scriptsize{$50 \leq \epsilon < 100$}}
  \put (62,2.3) {\scriptsize{$100 \leq \epsilon < 200$}}
  \put (82,2.3) {\scriptsize{$\epsilon \geq 200$ mm}}
  \end{overpic}
  \caption{Reconstructions obtained using Redwood \cite{choi2015robust}, ORB-SLAM3 \cite{campos2021orb}, and the proposed method. Colors depict errors (distance to the closest ground truth point).}
  \label{fig:reconstructions}
\end{figure}
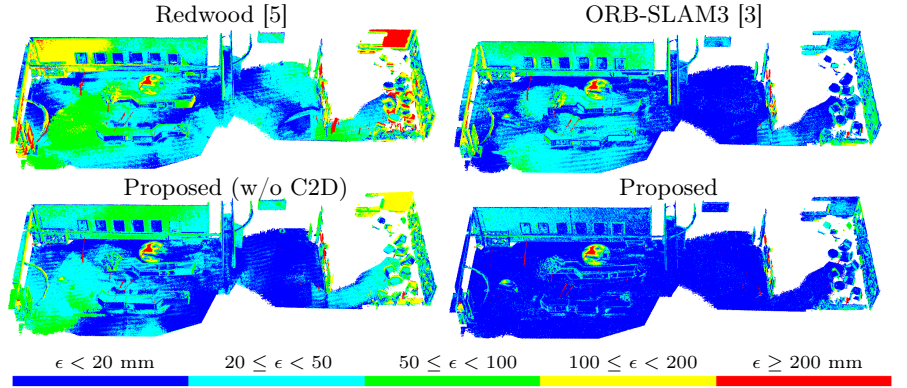

\subsection{Depth estimation}
We investigate whether the BS3D dataset can be used to train better models for monocular depth estimation. For this experiment, we use the state-of-the-art LeReS model \cite{Wei2021CVPR} based on ResNet50. The original model has been trained using 354k samples taken from various datasets \cite{zamir2018taskonomy,niklaus20193d,kim2018deep,hua2020holopix50k,xian2020structure}. We finetune the model using 16.5k samples from BS3D. We set the learning rate to 2e-5 and train only 4 epochs to avoid overfitting. Other training details, including loss functions are the same as in \cite{Wei2021CVPR}.

For testing, we use NYUD-v2 \cite{Silberman:ECCV12} and iBims-1 \cite{koch2018evaluation} that are not seen during training. We also evaluate using BS3D by sampling 535 images from an unseen part of the building. Table \ref{tab:depth} shows that finetuning improves the performance on iBims-1 and BS3D. The finetuned model performs marginally worse on NYUD-v2 which is not surprising considering that NYUD-v2 mainly contains room-scale scenes that are not present in BS3D. The qualitative comparison in Fig. \ref{fig:depth} also shows a clear improvement over the pretrained model on iBims-1 that contains both small and large scenes. The model trained only using BS3D cannot compete with other models, except on BS3D on which the performance is surprisingly good. The poor performance on other datasets is not surprising because of the small training set.

\begin{table*}
\newcolumntype{.}{D{.}{.}{-1}}
\centering
\setlength{\tabcolsep}{8.0pt}
\caption{Monocular depth estimation using LeReS \cite{Wei2021CVPR} trained from scratch using BS3D, pretrained model, and finetuned model. NUYD-v2 \cite{Silberman:ECCV12}, iBims-1 \cite{koch2018evaluation}, and BS3D are used for testing.}
\begin{tabular}{lllllll}
\toprule
\multicolumn{1}{c}{} &
\multicolumn{2}{c}{NYUD-v2 \cite{Silberman:ECCV12}} &
\multicolumn{2}{c}{iBims-1 \cite{koch2018evaluation}} &
\multicolumn{2}{c}{BS3D} \\
\cmidrule(r){2-3}
\cmidrule(r){4-5}
\cmidrule(r){6-7}
\multicolumn{1}{l}{Training data} & 
\multicolumn{1}{l}{AbsRel $\downarrow$} &
\multicolumn{1}{l}{$\delta_1$ $\uparrow$} &
\multicolumn{1}{l}{AbsRel $\downarrow$} &
\multicolumn{1}{l}{$\delta_1$ $\uparrow$} &
\multicolumn{1}{l}{AbsRel $\downarrow$} &
\multicolumn{1}{l}{$\delta_1$ $\uparrow$} \\
\cmidrule(r){1-7}
BS3D & 0.181 & 0.764 & 0.188 & 0.763 & 0.144 & 0.828 \\
Pretrained & \textbf{0.096} & \textbf{0.913} & 0.115 & 0.890 & 0.157 & 0.785 \\
Pre. + BS3D & 0.100 & 0.907 & \textbf{0.098} & \textbf{0.901} & \textbf{0.115} & \textbf{0.881} \\
\bottomrule
\end{tabular}
\label{tab:depth}
\end{table*}

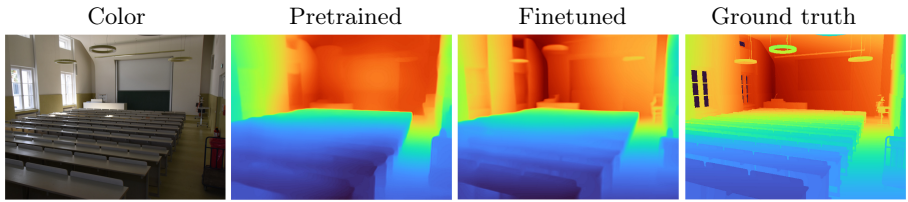
\begin{figure}
  \centering
  \begin{overpic}[width=1.00\textwidth]{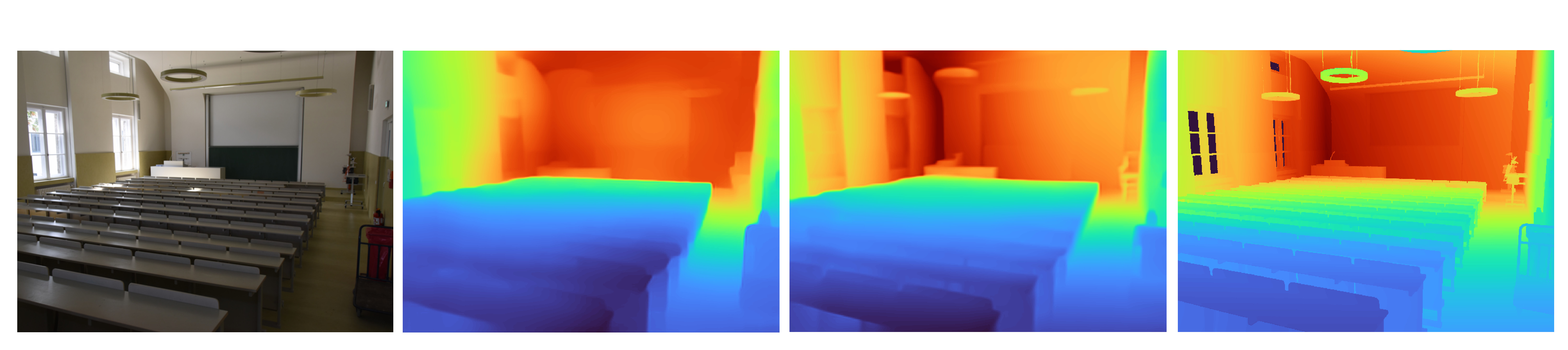}
  \put (10,20.3) {Color}
  \put (32,20.3) {Pretrained}
  \put (57,20.3) {Finetuned}
  \put (78,20.3) {Ground truth}
  \end{overpic}
  \caption{Comparison of pretrained and finetuned (BS3D) monocular depth estimation model LeReS \cite{Wei2021CVPR} on an independent iBims-1 \cite{koch2018evaluation} dataset unseen during training.}
  \label{fig:depth}
\end{figure}

\subsection{Visual-inertial odometry}
The BS3D dataset includes active infrared images along with color and IMU data. This opens interesting possibilities, for example, the comparison of color and infrared as inputs for visual-inertial odometry. Infrared-inertial odometry is an attractive approach in the sense that it does not require external light, meaning it would work in completely dark environments.

We evaluate OpenVINS \cite{geneva2020openvins}, ORB-SLAM3 \cite{campos2021orb}, and DM-VIO \cite{von2022dm} using color-inertial and infrared-inertial inputs. Note that ORB-SLAM3 has an unfair advantage because it has a loop closure detector that cannot be disabled. In the case of infrared images, we apply a power law transformation ($\overline{I}=0.04 \cdot I^{0.6}$) to increase brightness. As supported by \cite{von2022dm}, we provide a mask of valid pixels to ignore black areas near the edges of the infrared images. 
We adjust the parameters related to feature detection when using infrared images with \cite{geneva2020openvins,campos2021orb}. We use the standard error metrics, namely absolute trajectory error (ATE) and relative pose error (RPE) which measures the drift per second. The methods are evaluated 5 times on each of the 10 sequences (Table \ref{tab:sequences}).

From the results in Table \ref{tab:odometry}, we can see that ORB-SLAM3 has the lowest ATE when evaluating color-inertial odometry, mainly because of loop closure detection. In most cases, ORB-SLAM3 and OpenVINS fail to initialize when using infrared images. We conclude that off-the-shelve feature detectors (FAST and ORB) are quite poor at detecting good features from infrared images. Interestingly, DM-VIO performs better when using infrared images instead of color which is likely due to the infrared camera's global shutter and wider FoV. This result reveals the great potential of using active infrared images for visual-inertial odometry and the need for new research.

\begin{table}
\centering
\setlength{\tabcolsep}{12.5pt}
\caption{Evaluation sequences used in the visual-inertial odometry experiment. Last column shows if the camera returns to the starting point (chance for a loop closure).}
\begin{tabular}{lcccc}
\hline
Sequence & Duration (s) & Length (m) & Dimensions (m) & Loop \\
\hline
cafeteria & 200 & 90.0 & 12.4 x 15.7 x 0.8 & \checkmark \\
central & 242 & 155.0 & 25.5 x 42.1 x 5.3 & \checkmark \\
dining & 192 & 109.2 & 33.8 x 25.0 x 5.5 & \checkmark \\
corridor & 174 & 77.6 & 31.1 x 4.7 x 2.4 & \checkmark \\
foobar & 75 & 37.1 & 5.4 x 14.4 x 0.6 & \checkmark \\
hub & 124 & 52.3 & 11.4 x 5.9 x 0.7 & - \\
juice & 103 & 42.7 & 6.3 x 8.6 x 0.5 & - \\
lounge & 222 & 94.2 & 14.4 x 10.3 x 1.1 & \checkmark \\
study & 87 & 40.0 & 5.6 x 9.8 x 0.6 & - \\
waiting & 139 & 60.1 & 9.8 x 6.7 x 0.9 & \checkmark \\
\bottomrule
\end{tabular}
\label{tab:sequences}
\end{table}



\begin{table}
\centering
\setlength{\tabcolsep}{1.6pt}
\caption{Comparison of visual-inertial odometry methods using color-inertial and infrared-inertial inputs. Average absolute trajectory error (ATE) and relative pose error (RPE). Last column shows the percentage of successful runs.}
\begin{tabular}{l|cccc|cccc}
\toprule
\multicolumn{1}{c|}{} &
\multicolumn{4}{c|}{Color-inertial odometry} &
\multicolumn{4}{c}{Infrared-inertial odometry} \\
\cmidrule(r){2-5}
\cmidrule(r){6-9}
\multicolumn{1}{l|}{Method} & 
\multicolumn{1}{c}{\begin{tabular}{c}{ATE $\downarrow$}\\{(m)}\end{tabular}} &
\multicolumn{1}{c}{\begin{tabular}{c}{RPE $\downarrow$}\\{(deg/s)}\end{tabular}} &
\multicolumn{1}{c}{\begin{tabular}{c}{RPE $\downarrow$}\\{(m/s)}\end{tabular}} &
\multicolumn{1}{c|}{\begin{tabular}{c}{Succ. $\uparrow$}\\{(\%)}\end{tabular}} &
\multicolumn{1}{c}{\begin{tabular}{c}{ATE $\downarrow$}\\{(m)}\end{tabular}} &
\multicolumn{1}{c}{\begin{tabular}{c}{RPE $\downarrow$}\\{(deg/s)}\end{tabular}} &
\multicolumn{1}{c}{\begin{tabular}{c}{RPE $\downarrow$}\\{(m/s)}\end{tabular}} &
\multicolumn{1}{c}{\begin{tabular}{c}{Succ. $\uparrow$}\\{(\%)}\end{tabular}} \\
\cmidrule(r){1-9}
OpenVINS \cite{geneva2020openvins} & 0.347 & 0.37 & 0.031 & 76.0 & 0.597 & 0.42 & 0.057 & 36.0 \\
ORB-SLAM3 \cite{campos2021orb} & 0.298 & 0.29 & 0.026 & 100.0 & 0.193 & 0.29 & 0.025 & 24.0  \\
DM-VIO \cite{von2022dm} & 0.491 & 0.29 & 0.033 & 100.0 & 0.433 & 0.29 & 0.025 & 100.0 \\
\bottomrule
\end{tabular}
\label{tab:odometry}
\end{table}

\section{Conclusion}
We presented a framework for acquiring high-quality 3D reconstructions using a consumer depth camera. The ability to produce building-scale reconstructions is a significant improvement over existing methods that are limited to smaller environments such as rooms or apartments. The proposed C2D alignment enables the use of raw depth maps, resulting in more accurate 3D reconstructions. Our approach is fast, easy to use, and requires no expensive hardware, making it ideal for crowd-sourced data collection. We acquire building-scale 3D dataset (BS3D) and demonstrate its value for monocular depth estimation. BS3D is unique also because it includes active infrared images, which are often missing in other datasets. We employ infrared images for visual-inertial odometry, discovering a promising new research direction.

%
%
%
\bibliographystyle{splncs04}
\bibliography{bibliography}
\end{document}